# Deep Learning Method to Predict Wound Healing Progress Based on Collagen Fibers in Wound Tissue

Juan He[a,†], Xiaoyan Wang[b,†], Long Chen[a,*] , Yunpeng Cai[c,*] , Zhengshan Wang[a]

[a]Department of Computer and Information Science, Faculty of Science and Technology, University of Macau, Macau, 999078, China

[b]Institute of Translational Medicine, Faculty of Health Sciences & Ministry of Education Frontiers Science Center for Precision Oncology, University of Macau, Taipa, Macau, China

[c]Shenzhen Institute of Advanced Technology, Chinese Academy of Sciences, Shenzhen 518055, China

yc27967@umac.mo (Juan He), xiaoyanwang@um.edu.mo, longchen@umac.mo, yp.cai@siat.ac.cn, yc27436@umac.mo (Zhengshan Wang)

**Abstract:**

Wound healing is a complex process involving changes in collagen fibers. Accurate monitoring of these changes is crucial for assessing the progress of wound healing and has significant implications for guiding clinical treatment strategies and drug screening. However, traditional quantitative analysis methods focus on spatial characteristics such as collagen fiber alignment and variance, lacking threshold standards to differentiate between different stages of wound healing. To address this issue, we propose an innovative approach based on deep learning to predict the progression of wound healing by analyzing collagen fiber features in histological images of wound tissue. Leveraging the unique learning capabilities of deep learning models, our approach captures the feature variations of collagen fibers in histological images from different categories and classifies them into various stages of wound healing. To overcome the limited availability of histological image data, we employ a transfer learning strategy. Specifically, we fine-tune a VGG16 model pretrained on the ImageNet dataset to adapt it to the classification task of histological images of wounds. Through this process, our model achieves 82% accuracy in classifying six stages of wound healing. Furthermore, to enhance the interpretability of the model, we employ a class activation mapping technique called LayerCAM. LayerCAM reveals the image regions on which the model relies when making predictions, providing transparency to the model's decision-making process. This visualization not only helps us understand how the model identifies and evaluates collagen fiber features but also enhances trust in the model's prediction results. To the best of our knowledge, our proposed model is the first deep learning-based classification model used for predicting wound healing stages. Our proposed deep learning model not only accurately predicts wound recovery but also provides decision explanations and visualizations, which are of great value in clinical diagnosis and analysis of wound healing progression. The application of this approach holds the potential to bring more precise and effective medical practices to the field of wound treatment and management.

## 1 Introduction

During the skin wound healing process, the dynamic changes of collagen fibers play a central biological role. Precise monitoring of these changes is crucial for assessing the progress of wound healing and provides essential guidance for the formulation of clinical treatment strategies and drug screening[1]. However, traditional quantitative analysis methods focus solely on measuring spatial attributes such as the density and orientation of collagen fibers in histological images of wounds[2]. These quantitative methods lack clear threshold standards, making them challenging to use as a gold standard for classifying wounds at different stages of healing, thus presenting a challenge in accurately predicting the progression of wound healing. The rise of deep learning technology offers a revolutionary solution to this problem. By training deep neural networks, it is possible to automatically identify and analyze complex patterns in images, which shows great potential and promise in capturing subtle features of collagen fibers in histological images of wound tissue, leading to improved accuracy in predicting the progression of wound healing.

Histological images of wound tissue typically fall into two categories: invasive imaging and non-invasive imaging. Invasive imaging involves the process of cutting tissue samples into thin sections and then examining them under a microscope after applying specific stains to highlight collagen and its associated components. Common staining techniques used for this purpose include Masson's trichrome (MT) and Hematoxylin and Eosin (H&E) [3,4].On the other hand, non-invasive imaging refers to methods of acquiring images without the need to destroy tissue samples. Second Harmonic Generation (SHG) microscopy is a popular tool for non-invasive visualization of collagen fiber architecture [5]. SHG microscopy takes advantage of the nonlinear optical properties of collagen fibers to generate high-resolution images of the collagen structure.

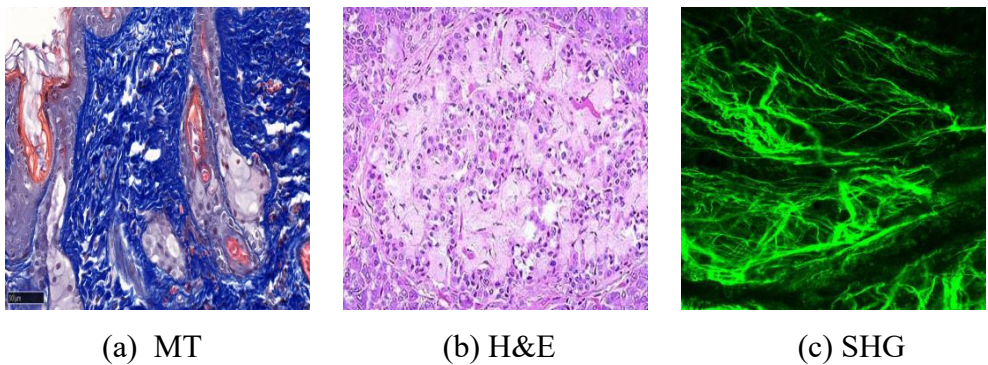

(a) MT   (b) H&E   (c) SHG

Figure1 Histological images of wound tissue

In the analysis of histological images from wound tissue, traditional analysis methods quantify collagen's spatial structure, aiding wound tissue identification and characterization[6-9]. Liu et al. devised a computational approach using curve transformation for fibrous collagen quantification[7,8]. Quin et al.'s study on MT-stained images showed scar tissue has denser collagen fibers and less directional variance than normal tissue[6]. Clemons introduced a method for collagen orientation assessment, revealing differences in coherency between normal and scar tissue [9]. It is important to highlight that while current research has shown significant differences in collagen spatial orientation and density between normal and wound tissues, there is a lack of significant differences among different stages of wound healing[6]. Consequently, it becomes challenging to determine pathological threshold values related to collagen characteristics for different stages of wound healing. This implies that solely relying on collagen density and directional variance to predict the extent of wound healing is insufficient when analyzing newly obtained histological images from wound tissue. As a result, the ability to provide precise diagnoses for cases of delayed wound healing is greatly limited in clinical practice.

In recent years, the application of deep learning in automated image analysis has gained widespread popularity, demonstrating remarkable advancements in the field of biomedical research[10,11]. Deep learning plays a pivotal role in tasks such as medical image analysis[12], disease diagnosis[13], and drug screening[14], driving significant breakthroughs and innovations in the healthcare industry. With regards to the specific task of evaluating wound healing based on histological images, deep learning models exhibit immense potential by leveraging their unique learning mechanisms to accurately capture the variations among different categories of images.

Currently, several deep learning models have achieved high-precision classification of normal and scar tissues, enabling the generation of collagen-positive maps based on extracted features, providing valuable insights into collagen fiber representation[15,16]. However, these models have encountered limitations in terms of their feature selection strategies, which have impeded the optimal extraction of collagen information. Additionally, as far as we are aware, there is currently a dearth of research utilizing deep learning to capture the dynamic variations of collagen at different stages of the wound healing process. This underscores the importance of further investigation to enhance our understanding of how collagen evolves during wound healing. Addressing this research gap will contribute to advancing our knowledge and could potentially lead to more comprehensive and accurate assessments of wound healing progression.

However, due to the limited availability of histological images from wound tissue in the real world, it is challenging to directly train a powerful deep learning model. This is because deep learning models typically require a large amount of data for training and parameter optimization, which may not be feasible for the target task. In such cases, the performance of the deep learning model is hindered. To overcome this challenge, transfer learning has emerged as an effective approach. Transfer learning leverages knowledge gained from related tasks that have already been learned, thereby accelerating the learning process for a new task [17]. Specifically, it involves training a powerful model on a similar task and then fine-tuning the model to adapt it to the target task [18]. By utilizing transfer learning, the model can leverage existing knowledge and overcome the limitations of scarce data for the target task.

In this paper, we apply the pre-trained VGG16 deep learning model, which was initially trained on the ImageNet dataset[20], to achieve the classification of wound healing progress through a process known as fine-tuning. The pre-trained VGG16 model has learned rich feature representations on ImageNet, and through transfer learning, we can leverage these generic features to accelerate the learning process for the classification of wound healing progress.

To enhance the interpretability of the model, we also introduce a visual interpretability technique called Class Activation Mapping (CAM)[21,22]. CAM technique identifies the specific areas that contribute most significantly to the model's predictions by weighting and fusing the model's features, revealing the decision basis of the model. Through the strategic weighting and integration of model features, the CAM technique effectively activates the regions corresponding to collagen fibers in histological images of wound tissue. These regions are pivotal for the accurate classification of wound healing stages. CAM offers a novel vantage point, enabling the visual inspection and analysis of collagen fibers' salient attributes within the tissue. This not only deepens our comprehension of the model's decision-making mechanisms but also underscores the significance of collagen fibers in the wound healing journey. By employing CAM, we can pinpoint the specific collagen fiber areas the model deems crucial for assessing progress in wound healing, thus casting light on their essential function in the overall process.

To the best of our knowledge, our proposed model is the first deep learning-based classification model used for predicting wound healing stages. The introduced deep learning model not only accurately predicts the extent of wound healing but also provides explanations and visualizations, which are highly beneficial for understanding the model's decision process, validating its accuracy, and ultimately assisting doctors or researchers in diagnosing and analyzing the progress of wound healing.

The main contributions of this paper are as follows:

(1) We collected histological images representing six categories of wound healing stages, including normal tissue from mice, wound tissue at day 0, day 3, day 7, and day 10 of healing, as well as wound tissue at day 10 with delayed healing due to diabetes in mice.

(2) We constructed a six-class classification model for predicting wound healing stages by fine-tuning a pre-trained VGG16 model.

(3) We introduced a visual interpretability method called LayerCAM, which highlights specific regions in histological images that contribute the most to the model's predictions, particularly the collagen fiber regions.

(4) We utilized traditional quantitative analysis methods to calculate the directional coherency of collagen fibers in the six categories of histological images and performed t-tests to assess the significant differences in inter-class coherency.

## 2. Related Work
### 2.1 Collagen in wound healing

Collagen is pivotal in wound healing, offering mechanical reinforcement, fostering cell movement and division, modulating cellular communication, and aiding in neovascularization [1,23].

The synthesis, organization, and restoration of collagen are critical to the healing cascade. Post-injury, collagen is produced and secreted by activated fibroblasts and myofibroblasts, which then accumulate around the wound site, forming robust fiber networks through cross-linking. This network supplies structural integrity, safeguards the wound, and serves as a platform for cellular movement [24,25].

Additionally, collagen influences cell migration and proliferation [26,27,28]. Its presence enhances cell movement within the wound and stimulates the growth of fibroblasts in adjacent tissues. This activity contributes to wound closure and the generation of new tissue. Collagen also interacts with cellular signaling pathways to regulate cellular responses and inflammatory processes during healing [29,30,31].

Neovascularization is a vital component of the healing process, and collagen serves as a supportive framework for this process, offering both physical support and directional cues for new blood vessel development [32,33]. Bioactive peptides within collagen are also implicated in the direct regulation of angiogenesis [34,35,36].

The strategic manipulation of collagen synthesis, organization, and repair can expedite wound closure and tissue regeneration. Building upon this understanding, we have developed a deep learning-based collagen image classification model that aims to evaluate the progression of wound healing. This model analyzes and classifies collagen images, providing valuable insights into the healing process.

**2.2 Deep Learning for Collagen Analysis**

This section delves into the progress made in utilizing deep learning models for the analysis and characterization of collagen structures in biological tissues. Through training a deep learning model, researchers have achieved remarkable results in predicting the elastic mechanical properties of tissues by analyzing second harmonic generation (SHG) images of collagen networks[37]. Furthermore, neural networks have been employed to automate the classification and segmentation of local collagen fiber orientations in both 2D and 3D images, offering valuable insights into the structure of collagen[38].

In the field of cancer research, the combination of ridge filters and deep neural networks has facilitated the development of powerful classifiers for collagen fibers based on breast cancer images, shedding light on the relationship between collagen structure and tumor progression[39]. Convolutional neural networks have also been utilized to analyze pathological images of breast cancer, uncovering a strong correlation between tissue stiffness and the presence of straight collagen fibers[40]. Additionally, the residual network model has been employed to create a classifier called collagenDL, accurately forecasting disease-free survival and overall survival in patients with stage II-III colon cancer[41].

Moreover, researchers have proposed convolutional neural network models for the classification of histology images of burn-induced scar tissue, enabling precise differentiation between normal and scar tissue, along with the quantitative assessment of collagen microstructures [15,16]. The U-Net model, specifically trained by Alan E. Woessner et al., stands out for its ability to accurately segment collagen-positive pixels, even in cases with varying imaging depths. This capability proves valuable in the quantitative application of second harmonic generation imaging in thick tissues[42]. Additionally, the U-Net deep learning technique has been successfully applied by Nicole Riberti et al. to characterize the intricate microstructure of collagen bundles surrounding dental implants, providing precise segmentation and analysis based on specific directional properties[43].

Furthermore, Park H and colleagues have proposed a method that harnesses the power of deep neural networks to extract precise collagen fiber centerlines from microscopy-based collagen images collected from pathological tissue samples. This method enables comprehensive quantitative measurements of fiber orientation, alignment, density, and length, yielding valuable insights into collagen structure[44].

Overall, the advancements in deep learning techniques have revolutionized the analysis and characterization of collagen structures, offering valuable insights into tissue properties and making significant contributions to fields such as cancer research and tissue engineering.

## 3 Materials and Methods
### 3.1 Data Information
#### 3.1.1 Animal Model Preparation

All animal experiments adhered to a protocol sanctioned by the University of Macau's Sub-panel on Animal Research Ethics （UMARE – 019 - 2022）. Male nude mice, aged 6-8 weeks and weighing between 20-23g, were sourced from the University of Macau's animal facility. The excisional wound splinting model was employed as delineated by X. Wang[45] (2019). In summary, each mouse was anesthetized before two full-thickness skin wounds of 5 mm diameter were inflicted with a biopsy punch on either side of the midline. A silicone splint encircled each wound to maintain its position while a 3M Tegaderm dressing provided coverage. The wounded skin was extracted for histological examination at specified intervals post-injury (e.g., days 0, 3, 7, 10).

For the purpose of conducting experiment on diabetic chronic wounds, transgenic mice (B6.Cg-Lepob/J) were utilized. Following mating and breeding, the hybrid Lepob mice were identified. The ob+ mice served as the control group, representing normal physiological conditions, while the ob/ob mice were considered diabetic. The skin wound model was implemented on both types of mice 12 weeks after breeding, using the same specific methodology as previously described.

#### 3.1.2 Histological Analysis Using Masson's Trichrome

Tissue specimens were fixed in 4% paraformaldehyde, dehydrated in a graded ethanol series and embedded in paraffin. For direct visualization of collagen fibres and histological assessment of collagen deposition, Masson trichrome staining was performed using the Kit (Heart Biological Technology). Optical photomicrographs were obtained at 40× magnification (Hamamatsu NanoZoomer S60, 0.26 μm/pixel resolution) with a digital camera, using a consistent setting. In the histology images, the collagen fibers are mostly stained in blue whereas the appendages, such as hair follicles, sweat glands, and sebaceous glands, are stained in red or purple.

#### 3.1.3 Data Processing

We collected three datasets of MT-stained histological images representing six categories of wound healing stages, including normal tissue from mice, and wound tissues from mice at the healing stages of day 0, day 3, day 7, and day 10, as well as wound tissues from mice with delayed healing due to diabetes on day 10. Table 1 presents the statistical overview of the three datasets.

Table1 Data distributions

|  | Control | Day0 | Day3 | Day7 | Day10 | Delay Day10 |
|---|---|---|---|---|---|---|
| Dataset1 | 28 | 12 | 15 | 31 | 38 | - |
| Dataset2 | 21 | - | - | - | 28 | 33 |
| Dataset3 | 23 | - | - | - | 44 | 33 |
| **Sum** | **72** | **12** | **15** | **31** | **110** | **66** |

To prepare the data for training, validation, and testing of the deep learning model, we merged three datasets and divided them into groups using a ratio of 6:2:2. For the training set, we applied various data augmentation techniques such as flipping and rotating to increase the diversity of the data. Since the distribution of the images among the groups was imbalanced, we employed oversampling to balance the training set. This oversampling technique ensured that each of the six classes had an equal representation in the final training set.

To ensure compatibility with the VGG16 CNN model, which requires input images of size 224x224, we used bilinear interpolation to resize all the images in the dataset to the desired dimensions. The specific breakdown of the dataset is show as Table2.

Table2 Data split

| Classes | Training | Validation | Testing | Augmented Training Data | Balanced Training Data |
| --- | --- | --- | --- | --- | --- |
| Control | 44 | 15 | 13 | 528 | 792 |
| Day 0 | 8 | 2 | 2 | 96 | 792 |
| Day 3 | 10 | 3 | 2 | 120 | 792 |
| Day 7 | 19 | 6 | 6 | 228 | 792 |
| Day10 | 66 | 22 | 22 | 792 | 792 |
| Delay Day10 | 40 | 13 | 13 | 480 | 792 |

## 3.2 Network Architecture and Training
### 3.2.1 Pre-Trained CNN Architecture

VGG16 is a widely-adopted convolutional neural network (CNN) architecture, which comprises 16 layers, including 13 convolutional layers and 3 fully connected layers. In this research, the pre-trained VGG16 model, initially trained on the ImageNet dataset, was utilized as the base architecture. Fine-tuning was conducted to adapt the model specifically for wound histological images classification.

By leveraging transfer learning, the knowledge and learned features from the ImageNet dataset can be effectively transferred to the task of classifying wound histological images. This approach reduces the need for training the entire model from scratch and allows the model to benefit from the pre-existing knowledge. The architecture of our proposed method for classifying wound histological images is depicted in Figure 2.

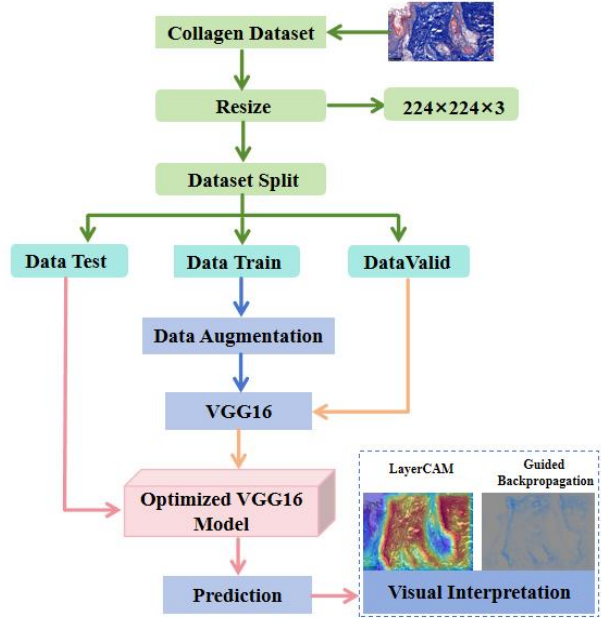

Figure2 Global overview of the proposed method

**3.2.2 Visual Interpretation**

LayerCAM is an extension of CAM that generates more detailed visualizations by producing class activation maps at multiple layers of the CNN[22]. The fundamental concept behind LayerCAM involves utilizing backward class-specific gradients to assign distinct weights to each spatial location in a feature map. The computation of gradients is shown in Equation (1), where $y_c$ is the predicted score for input image belonging to class *c*. *A* represents all the feature maps outputted by a specific layer of the model. $A^k_{ij}$ denotes the feature value at the *i-th* row and *j-th* column of the *k-th* feature map, while $g^{kc}_{ij}$ represents the gradient corresponding to that feature value.

$$g^{kc}_{ij} = \frac{\partial y^c}{\partial A^k_{ij}} \tag{1}$$

Positive gradients at a specific location in the feature map suggest that increasing the intensity of that location would positively influence the prediction score for the target class. In such cases, the corresponding gradients are used as weights. Conversely, locations with negative gradients are assigned a weight of zero, as intensifying them would negatively impact the prediction score.

The ReLU operation described in Equation (2) selectively preserves positive gradient values, while setting negative gradient values to 0. In Equation (3), $m_{ij}^k$ represents the class activation mapping value of $A^k_{ij}$. Finally, the class activation mapping $M^c$ is obtained by performing a channel-wise sum of $m^k$, and the ReLU function is applied to retain positive activation values. By leveraging the LayerCAM technique, we are able to locate collagen regions from the layers of the model.

$$ReLU(g^{kc}_{ij}) = max(0, g^{kc}_{ij}) \tag{2}$$

$$m_{ij}^k = ReLU\left(g_{ij}^{kc}\right) \cdot A_{ij}^k \tag{3}$$

$$M^c = ReLU \sum_k m^k \tag{4}$$

Guided Backpropagation is a technique used to visualize and interpret the decision-making process of a convolutional neural network (CNN). Its purpose is to highlight the influential regions of an input image that have the most significant impact on the network's output prediction. And the Guided Backpropagation method selectively propagates gradients during where both the corresponding feature value and gradient value are positive, setting the gradient values to 0 for other positions. Compared to LayerCAM, the Guided Backpropagation can retains the fine-grained information of crucial pixels for the prediction. However, it relatively weak in distinguishing subtle differences between different classes, lacking sensitivity to class-specific information and explicit discriminative ability and may experience a decrease in interpretability accuracy as it may incorrectly highlight regions unrelated to the target class or portray them as important features.

To achieve more interpretable and comprehensive visualization results, we leverage the strengths of Guided Backpropagation and LayerCAM. We perform element-wise multiplication between the LayerCAM heatmap and the positive gradients computed by Guided Backpropagation to obtain the visualization result. Figure 3 illustrates the mechanism of collagen visualization. This approach allows us to visualize the network's predictions for wound healing in a more comprehensive manner, taking advantage of accurate collagen localization and preserving fine-grained details. By combining these techniques, we enhance the interpretability and richness of the visualizations.

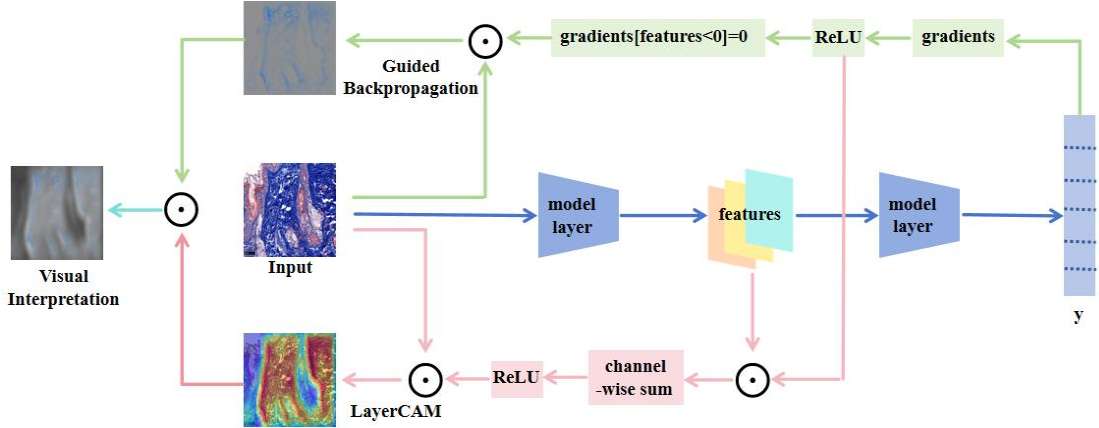

Figure3 Visualization framework

## 4 Results and Discussion
### 4.1 Network Training and Visual Interpretability

During our network training, we employed transfer learning using the pre-trained VGG16 model trained on the ImageNet dataset. As our MT-stained histological images consist of six classes, we modified the classification layer of VGG16 to accommodate these six classes instead of the original 1000 classes. We set the learning rate to 0.0001, which controlled the rate at which the model updated its parameters during training, balancing between convergence speed and avoiding overshooting the optimal solution. The training was conducted for 40 epochs, with a batch size of 16. The model was trained on NVIDIA Corporation Device 2230 GPU, using Python version 3.8.0, and the PyTorch deep learning framework version 1.8.1.

To evaluate the performance of the model, we used two metrics: AUC (Area Under the Curve) and ACC (Accuracy). AUC provided insights into the model's ability to discriminate between the different classes of collagen images, while the ACC metric measured the overall accuracy of the model's predictions.

The six-class classification model achieved an accuracy of 82% and an area under the curve (AUC) of 97% on the test dataset. The accuracy for each class is shown in Table 3. And the ACC curve is depicted in Figure 4, the AUC curve is shown in Figure 5, and the loss curve is illustrated in Figure 6.

Additionally, we employed Guided Backpropagation and LayerCAM techniques to accurately localize the regions of interest (ROI) in the images. Figure 7 provides visual interpretability examples for various stages of wound healing. This analysis demonstrated that our model specifically focused on learning the collagen fibers of histological images for classifying the stages of wound healing, rather than other tissue structures such as cell nuclei. Moreover, this interpretability is facilitated through visual overlays and heatmaps that highlight the areas of the histological images where the model finds the most relevant features for its predictions. This not only aids in validating the model's accuracy but also provides insights into the underlying biological processes captured by the images. Consequently, our model serves not only as a predictive tool but also as an analytical one, contributing to a deeper understanding of wound healing dynamics and potentially guiding the development of more effective treatments for delayed healing conditions.

In summary, by leveraging advanced image analysis techniques of deep learning, our model can discern subtle changes in collagen fibers throughout the various stages of wound healing. This capability allows the model to accurately predict the progression of healing and identify any deviations from the norm, such as delayed healing. The visual interpretability aspect of our model provides a layer of transparency that enables researchers and clinicians to understand the decision-making process behind the model's predictions.

Table3 Six-class accuracy

|  | Control | Day0 | Day3 | Day7 | Day10 | Delay Day10 | Mean |
|---|---|---|---|---|---|---|---|
| ACC | 0.93 | 1.0 | 0.73 | 0.66 | 0.74 | **0.86** | 0.82 |

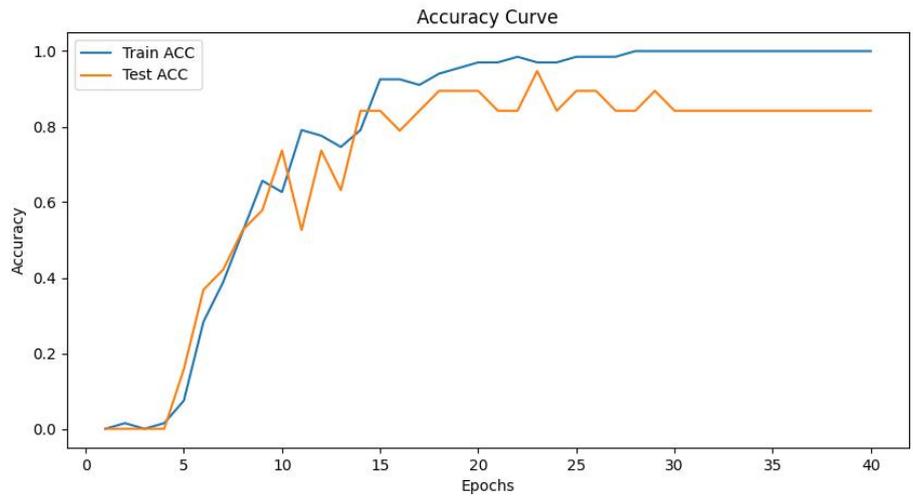

Figure4 Accuracy

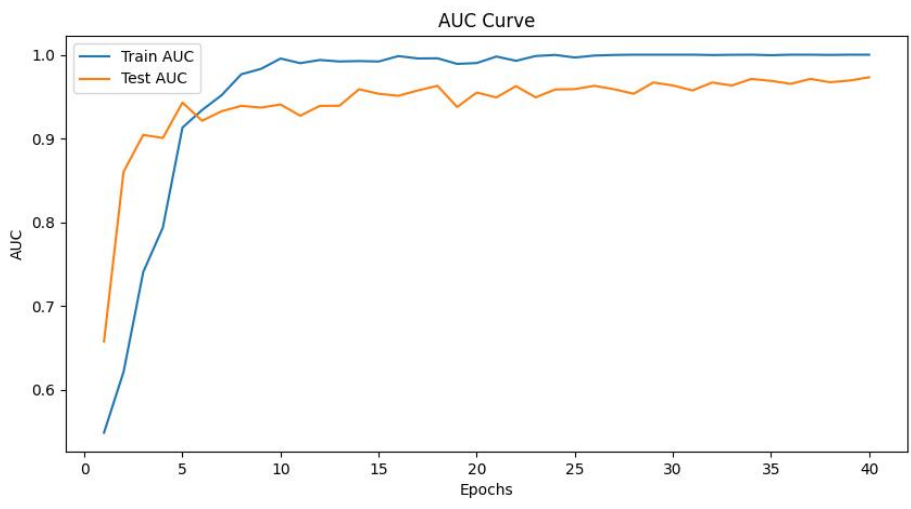

Figure5 AUC

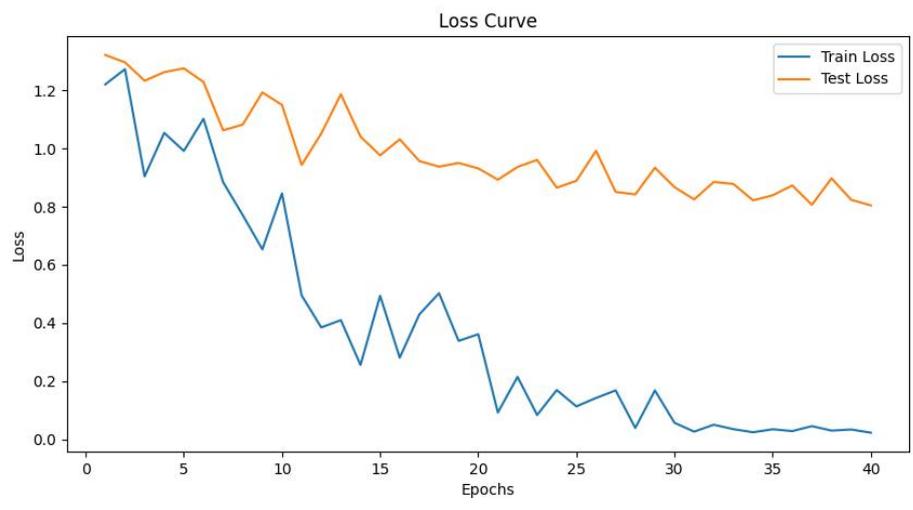

Figure6 Loss

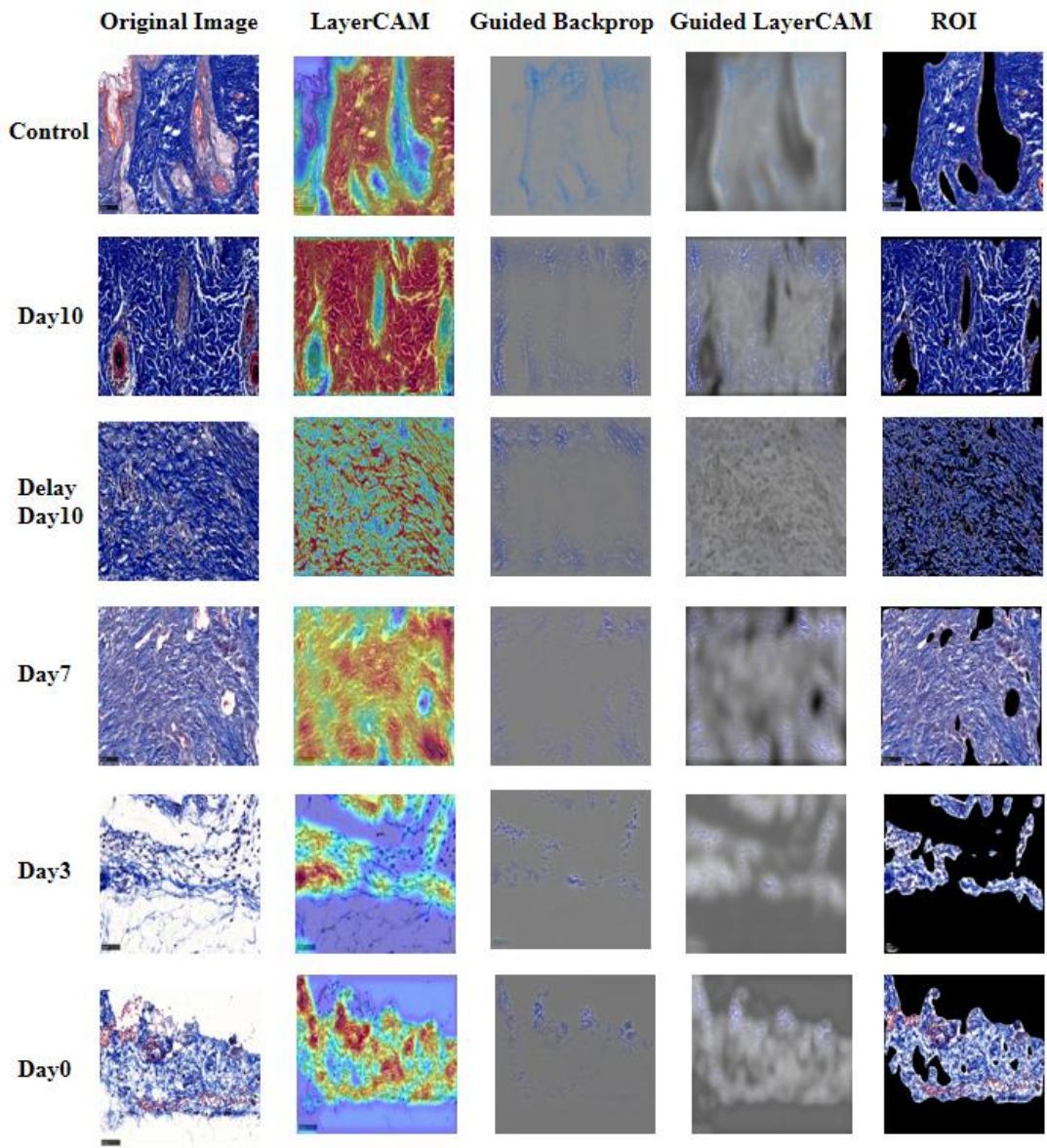

Figure 7 Visual Interpretability

### 4.2 Quantification of Collagen Fiber Orientation

The orientation of collagen fibers undergoes distinct variations during different stages of wound healing and plays a pivotal role in differentiating between these stages. In normal skin tissue, collagen fibers align in an organized manner, forming parallel fiber bundles. This arrangement contributes to a higher level of coherency[46]. When skin tissue is damaged, the alignment of collagen fibers in the wound tissue typically undergoes changes. During the initial stages of wound healing, the newly formed collagen fibers in the wound are relatively loose and irregular. They are shorter in length and have smaller diameters, resulting in lower coherency values. As the healing process progresses, collagen fibers gradually become more densely packed, with thicker diameters and longer lengths. They also exhibit a more regular arrangement, leading to increased coherency values[46]. Quantitative analysis of collagen fiber orientation offers a comprehensive understanding of the intricate details and underlying mechanisms of wound healing, ultimately guiding clinical practices. In this study, we utilized the coherency metric to precisely quantify and analyze the orientation of collagen fibers[46].

ImageJ is a versatile tool widely used for image processing and analysis in scientific research. Fiji (Fiji Is Just ImageJ) is an expanded version of ImageJ that offers enhanced functionality and additional plugins and tools [47]. In our analysis, we initially extracted collagen fibers, represented by the blue regions, from histological images based on chromatic saturation. Subsequently, we employed the OrientationJ plugin in Fiji to calculate the coherency of collagen fiber orientation.

We computed the coherency of collagen fiber orientation for the three sets of histological images listed in Table 1. The distribution of coherency for each dataset is represented by the box plots shown in Figures 8 and 9. Average and median coherency values are provided in Tables 4, 6, and 8. The aforementioned figures reveal that the directionality consistency of collagen fibers gradually increases during the healing process. In the early stages of wound healing, the coherency of collagen fibers in the wound tissue is lower compared to that in normal tissue. However, in the later stages of wound healing, such as on the 10th day of wound healing, the coherency of collagen fibers in the wound tissue surpasses that of normal tissue. Additionally, delayed wound healing tissue exhibits higher coherency of collagen fibers compared to non-delayed wound healing tissue.

We also conducted t-tests to calculate the significance differences in collagen fiber coherency among different categories of histological images within three datasets listed in Table 1, as shown in Tables 5, 7, and 9. Table 4 indicates that in the first dataset, there are significant differences in the direction coherency of collagen fibers between histological images at Day 0 of the healing stage and normal tissue, as well as between histological images of wound tissue at Day 7 and wound tissue at Day 0. Moreover, there are significant differences in the direction coherency of collagen fibers between histological images of wound tissue at Day 10 and all other categories of histological images. Table 6 shows that in the second dataset, there are significant differences in the direction coherency of collagen fibers between histological images of delayed wound healing tissue and normal tissue, as well as between histological images of delayed wound healing tissue and histological images of wound tissue at Day 10. Table 8 reveals that in the third dataset, there is a significant difference in the direction coherency of collagen fibers between histological images of delayed wound healing tissue and histological images of wound tissue at Day 10. The above analysis suggests that significant differences in the direction coherency of collagen fibers among different categories of histological images may not always be present. This indicates that traditional methods of quantifying collagen fibers, such as directional coherency, have limitations in distinguishing different stages of wound healing.

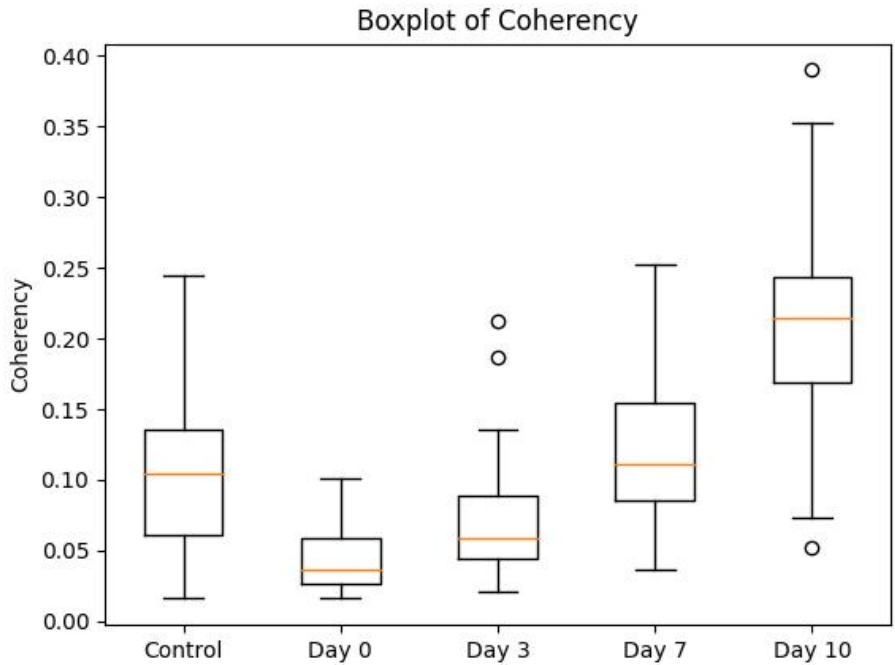

Figure8 Coherency of the first datasets

Table4 Coherency of the first datasets

|  | Control | Day0 | Day3 | Day7 | Day10 |
|---|---|---|---|---|---|
| Number of Images | 28 | 12 | 15 | 31 | 38 |
| Mean Coherency | 0.106 | 0.046 | 0.08 | 0.126 | 0.21 |
| Median Coherency | 0.104 | 0.0365 | 0.059 | 0.111 | 0.215 |

Table5 The p-value of the t-test for coherency of the first datasets

|  | Control | Day0 | Day3 | Day7 | Day10 |
|---|---|---|---|---|---|
| Control | - | **0.0005** | 0.1320 | 0.1651 | **0.0000** |
| Day0 | **0.0005** | - | 0.0707 | **0.0000** | **0.0000** |
| Day3 | 0.1320 | 0.0707 | - | **0.0132** | **0.0000** |
| Day7 | 0.1651 | **0.0000** | **0.0132** | - | **0.0000** |
| Day10 | **0.0000** | **0.0000** | **0.0000** | **0.0000** | - |

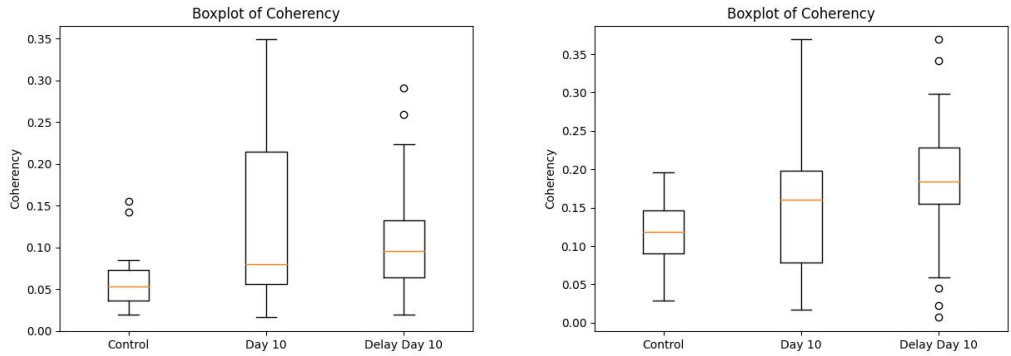

(a) Coherency of the second datasets     (b) Coherency of the third datasets

Fig9 Coherency of the two delay datasets

Table6 Coherency of the second datasets

|  | Control | Day10 | Delay day 10 |
|---|---|---|---|
| Number of Images | 21 | 28 | 33 |
| Mean Coherency | 0.0604 | 0.1273 | 0.1068 |
| Median Coherency | 0.053 | 0.08 | 0.096 |

Table7 The p-value of the t-test for coherency of the second datasets

|  | Control | Day 10 | Delay day 10 |
|---|---|---|---|
| Control | - | **0.0119** | **0.0042** |
| Day 10 | **0.0119** | - | 0.3941 |
| Delay day 10 | **0.0042** | 0.3941 | - |

Table8 Coherency of the third datasets

|  | Control | Day10 | Delay day 10 |
|---|---|---|---|
| Number of Images | 23 | 44 | 33 |
| Mean Coherency | 0.12 | 0.15125 | 0.18603 |
| Median Coherency | 0.118 | 0.16 | 0.184 |

Table9 The p-value of the t-test for coherency of the third datasets

|  | Control | Day 10 | Delay day 10 |
|---|---|---|---|
| Control | - | 0.1054 | **0.0010** |
| Day 10 | 0.1054 | - | 0.0760 |
| Delay day 10 | **0.0010** | 0.0760 | - |

**4.3 Discussion**

    Through detailed quantitative analysis of collagen fiber direction, we have observed that as the wound healing process progresses, the arrangement of collagen fibers in the wound tissue gradually transitions from disorganized and chaotic to a more orderly and coherent pattern. This increasing coherency signifies an important transformation in the structural reconstruction and functional recovery during wound healing. The ordered alignment of collagen fibers is crucial for providing sufficient

mechanical strength and tissue support, serving as a significant indicator of healing quality.

However, it is worth noting that the changes in collagen fiber coherency between different stages of wound healing may not always be significant. These subtle differences can be challenging to accurately capture using traditional quantitative analysis methods. Particularly when comparing different stages of wound healing or analyzing cases with delayed healing due to certain pathological conditions (such as diabetes), traditional methods may not effectively reveal the subtle variations in collagen fiber arrangement. This challenge highlights the limitations of traditional analysis methods in terms of accuracy and sensitivity. They may not fully identify and quantify the subtle changes in collagen fiber alignment, which are crucial for precisely determining the specific stages of wound healing and evaluating treatment outcomes. Therefore, the reliability of these traditional methods as standards for distinguishing between different stages of wound healing may be questioned.

In contrast, our proposed deep learning-based classification method for assessing the degree of wound healing leverages its advanced capabilities in image recognition and feature extraction. It provides a more precise and sensitive means of identification. This method is particularly effective in revealing subtle changes that may be overlooked by traditional statistical methods, especially in detecting cases of delayed healing. For example, by inputting histological images of wound tissue samples exhibiting delayed healing on Day 10 into our deep learning model, the model can identify features inconsistent with normal healing processes and classify them as cases of delayed healing. Such real-time feedback is invaluable for healthcare professionals as it allows them to intervene promptly, adjust treatment plans, and facilitate smooth wound healing. This capability greatly complements the limitations of traditional statistical methods when analyzing complex biomedical data. Furthermore, our deep learning model provides visual interpretability by annotating regions of interest (ROIs) on histological images that the model recognizes, thereby making the decision-making process more transparent. This visualization technique not only helps researchers and physicians understand how the model makes classification predictions but also enhances the trustworthiness and reliability of the model's results.

Our proposed deep learning model can complement traditional quantitative methods. By combining traditional quantitative analysis methods for collagen fiber quantification with our deep learning-based classification technique for assessing the degree of wound healing, we can achieve a more comprehensive and detailed evaluation of wound tissue healing status. This multidimensional assessment approach empowers healthcare professionals with more accurate monitoring capabilities, enabling real-time tracking of wound healing progress and timely adjustment of treatment plans, particularly when dealing with complex or challenging-to-heal wounds.

**5 Conclusion**

In our research, we have proposed an innovative method that utilizes convolutional neural networks (CNN) to classify histological images and accurately evaluate the degree of wound healing. This approach has achieved accuracy rate of up to 82% when classifying six different stages of healing. To gain a deeper understanding of the model's decision-making process, we have employed advanced visualization techniques, such as class activation mapping (CAM) and guided backpropagation. These techniques allow us to precisely identify and visualize the regions of collagen fiber distribution in the images, shedding light on how the model utilizes these crucial features to make classification decisions.The visualization results not only enhance our understanding of the model's predictions but also increase its transparency and credibility. This intuitive presentation provides healthcare professionals with a valuable tool to better comprehend and assess the status of wound healing, especially when dealing with complex wounds that may exhibit delayed healing.

Looking towards the future, we plan to expand our research by incorporating second harmonic generation (SHG) collagen imaging technology. This will help improve the accuracy of analyzing collagen. SHG technology provides high-contrast images of collagen fibers without the need for staining, which is essential for studying the dynamic changes in collagen during the wound healing process. By incorporating this approach, we aim to achieve a more accurate assessment of the dynamics of wound healing, thereby providing stronger support for clinical treatments and paving the way for further research into the pathological mechanisms of delayed healing.